\def\year{2020}\relax
\documentclass[letterpaper]{article} 
\usepackage{aaai20}  
\usepackage{times}  
\usepackage{helvet} 
\usepackage{courier}  
\usepackage[hyphens]{url}  
\usepackage{graphicx} 
\usepackage{graphicx}  
\usepackage{graphicx}   
\usepackage{subfigure}
\usepackage{multirow}
\usepackage{booktabs}
\usepackage{diagbox}
\usepackage{bbm}
\usepackage{multicol}
\usepackage{blindtext}
\usepackage{amsmath,amsfonts,amssymb}
\usepackage{bm}
\usepackage{mathtools}
\usepackage{lipsum,booktabs}
\usepackage{url}
\usepackage{algorithm}
\usepackage{algpseudocode}

\DeclareMathOperator*{\bilstm}{Bi-LSTM}
\DeclareMathOperator*{\lstm}{LSTM}

\DeclareMathOperator*{\score}{Score}
\DeclareMathOperator*{\ffn}{FFN}
\DeclareMathOperator*{\relu}{ReLU}
\DeclareMathOperator*{\mlp}{MLP}
\DeclareMathOperator*{\sigmoid}{sigmoid}
\DeclareMathOperator*{\softmax}{softmax}
\DeclareMathOperator*{\Afind}{\textsc{Find}}
\DeclareMathOperator*{\Acount}{\textsc{Count}}
\DeclareMathOperator*{\Ain}{\textsc{In}}
\DeclareMathOperator*{\Aunion}{\textsc{Union}}
\DeclareMathOperator*{\Ainter}{\textsc{Inter}}
\DeclareMathOperator*{\Adiff}{\textsc{Diff}}
\DeclareMathOperator*{\Agreater}{\textsc{Greater}}
\DeclareMathOperator*{\Aless}{\textsc{Less}}
\DeclareMathOperator*{\Aequal}{\textsc{Equal}}
\DeclareMathOperator*{\Aargmax}{\textsc{Argmax}}
\DeclareMathOperator*{\Aargmin}{\textsc{Argmin}}

\DeclareMathOperator*{\Aset}{\textsc{Set}}
\graphicspath{{./figs/}}
\nocopyright
\title{Effective Search of Logical Forms for Weakly Supervised \\ Knowledge-Based Question Answering}
\author{
Tao Shen$^{1}$\thanks{~~~Work done while the author was an intern at Microsoft, Beijing, China. }, 
Xiubo Geng$^2$, Tao Qin$^2$, Guodong Long$^1$, Jing Jiang$^1$ and Daxin Jiang$^2$ \\
$^1$Centre for AI, School of Computer Science, FEIT, University of Technology Sydney \\
$^2$Microsoft, Beijing, China \\
{\tt tao.shen@student.uts.edu.au},~~{\tt \{guodong.long,jing.jiang\}@uts.edu.au} \\
{\tt \{xiubo.geng,taoqin,djiang\}@microsoft.com} \\
}
 
\begin{document}

\maketitle

\begin{abstract}
Many algorithms for Knowledge-Based Question Answering (KBQA) depend on semantic parsing, which translates a question to its logical form. When only weak supervision is provided, it is usually necessary to search valid logical forms for model training. However, a complex question typically involves a huge search space, which creates two main problems: 1) the solutions limited by computation time and memory usually reduce the success rate of the search, and 2) spurious logical forms in the search results degrade the quality of training data. These two problems lead to a poorly-trained semantic parsing model.  In this work, we propose an effective search method for weakly supervised KBQA based on operator prediction for questions. With search space constrained by predicted operators, sufficient search paths can be explored,  more valid logical forms can be derived, and operators possibly causing spurious logical forms can be avoided. As a result, a larger proportion of questions in a weakly supervised training set are equipped with logical forms, and fewer spurious logical forms are generated. Such high-quality training data directly contributes to a better semantic parsing model. Experimental results on one of the largest KBQA datasets (i.e., CSQA) verify the effectiveness of our approach: improving the precision from 67\% to 72\% and the recall from 67\% to 72\% in terms of the overall score. 
\end{abstract}


\section{Introduction} \label{sec:introduction}

Knowledge-based question answering (KBQA) interacts with a knowledge base (KB) and draws a correct answer for a factoid question. Many top-performing approaches to KBQA are based on a semantic parsing framework, that is, translating a natural language question into corresponding logical form in the light of pre-defined grammars~\cite{artzi2013weakly,vlachos2014new,suhr2018learning}.
For example, ``\textit{how many people have birthplace at Provence}'' has a corresponding logical form \textsc{Count(Find( {\textit{Provence}}, \textsl{{place-of-birth}}))}. The logical form is then executed by KB system to retrieve an answer.  

To train a semantic parser, ideal training example is in the format of $\langle${\it question, logical form}$\rangle$. However, it usually requires some expertise to compose logical forms, especially for complex questions~\cite{webquestions,wikitablequestions,csqa}. Therefore, it is not realistic to employ crowdsourcing to scale out the size of such training data. To circumvent this challenge, a weakly supervised training setting was proposed. The idea is to create training example in the format $\langle${\it question, answer}$\rangle$ instead of $\langle${\it question, logical form}$\rangle$, since it is easier to get the answer for a factoid question than writing the corresponding logical form. However, answers cannot be directly used to train a semantic parser. Therefore, given a factoid question, a crucial step in weak supervision is to automatically search for valid logical form over a knowledge base, which must lead to the given ground-truth answer after its execution. Logical forms derived from the searching process will then be considered as fully-supervised training targets for a semantic parser.

In this case, the quality of a semantic parser depends on the effectiveness of upstream searching process for logical forms. However, the search space for eligible logical forms can be very large~\cite{iyyer2017search}. For example, a complex question frequently involves 7 to 8 steps, and in each step an operator is chosen from up to 20 candidates. The size of search space is then about $20^7\sim 20^8$. Although we may leverage the constraint of grammars to prune the search space, it can still be at the magnitude of $10^4\sim 10^5$. The large search space results in two challenges as follows. 

First, it may not be practical to exhaustively search the whole hypothesis space, since this takes huge cost in computation time and memory to verify each candidate by executing the logical form in a large-scale KB. Therefore, a usual practice is to search a randomly-selected, middle-sized subspace. However, such incomplete search possibly misses valid logical forms. In our empirical study, we define {\em search success ratio} as the number of questions for which the subspace search can find valid logical forms, divided by the total number of questions. We applied some traditional search algorithms, such as naive BFS \cite{d2a}, on a public dataset CSQA \cite{csqa}, and found the success ratio is very low. For example, the search success ratio for \textit{comparative} and \textit{quantitative} questions are barely 25\% and 43\% respectively. In other words, for a large percentage of these questions, there are no corresponding logical forms generated as training data. As shown in our experiment part (Section \ref{sec:exp_qa}), the insufficient training data can negatively impact the performance of a semantic parser. 

Second, even if we could overcome the practical resource constraints and search the entire space, we are likely to find spurious logical forms rather than correct ones. A spurious logical form does not match the semantic meaning of the original question, but coincidentally results in ground-truth answer if executed over the KB. 
For example, for the question ``\textit{Which occupations do Camil Samson and Daniel von Moser do for a living}", 
due to the two persons have the same occupation, all following three logical forms, i.e., 
1) ``\textsc{Find(\textit{Camil Samson}, \textsl{{Occupation}})}'', 
2) ``\textsc{Find(\textit{Daniel von Moser}, \textsl{{Occupation}})}'', 
and 3) ``\textsc{Union(Find(\textit{Camil Samson}, \textsl{{Occupation}}), Find(\textit{Camil Samson}, \textsl{{Occupation}}))}'', 
lead to the ground-truth answer, but only the last one is correct. 
To measure the severity of spurious logical forms, we conducted a quantitative analysis over randomly held-out examples by human evaluation on CSQA, and found that up to 54.5\% of the search results were spurious. A large percentage of spurious forms in the training set can introduce high noise to diminish the performance of a semantic parser.

Several prior works have been proposed to reduce search space or decrease spurious logical forms, which can be categorized into two ways. 
First, some methods use techniques to reduce search space but still suffer from spurious logical forms, such as macro grammars \cite{macrogrammar} and logical form sketch \cite{dong2018coarse}. 
Second, under reinforcement learning, these works try to gradually reduce spurious logical forms while their model are iteratively trained on weakly supervised data, such as iterative search \cite{dasigi2019iterative},
but these methods still suffer from high failure ratio due to exponentially-growing search space, and are probably not suitable for supervised settings. 

In this work, we propose a novel approach to effectively search for logical forms over a large-scale knowledge base by introducing an operator predictor\footnote{An operator is an action unit we take when querying the KB, e.g. finding objects given a subject-predicate pair, counting the number of an entity set, comparing two numbers, etc.}. Intuitively, we can estimate operator candidates for a given question based on its semantics. For example, the phrase ``the most'' may suggest $\Aargmax$ and ``less than'' may suggest $\Aless$. With the constraint of the predicted operator set, searching for valid logical forms will result in a lower percentage of spurious logical forms and a higher ratio of the search success. In turn, high-quality training data will improve the accuracy of downstream question-to-logical-form translation model. Additionally, the predicted small set of operators can also be easily integrated into translation model's decoder to improve performance by providing constraints.

Experiments on CSQA dataset \cite{csqa}, one of the largest weakly supervised KBQA datasets over a large-scale KB with complex questions, verify the effectiveness of this approach. In particular, by searching logical forms with our approach, the percentage of spurious logical form is reduced from 55\% to 27\% by human evaluation, and the search success ratio increases from 71\% to 80\%; for KBQA task, the overall score is significantly improved compared to the baseline i.e., 5\% growth for both recall and precision.

\section{Our Approach}

This section starts with an introduction to grammars and logical form. Then, an outline of proposed approach and the implementation of models are elaborated. 

\begin{table}[t] \small
	\centering
	\begin{tabular}{l|l}
		\hline
		\textbf{Alias} & \textbf{Operator}  \\ \hline
		A1/A2/A3&            $start \rightarrow set/num/bool$          \\
		A4&          $set \rightarrow \Afind(set, p)$\\
		A5&          $num \rightarrow \Acount(set)$\\
		A6&          $bool \rightarrow \Ain(e, set)$\\
		A7/A8/A9&          $set \rightarrow \Aunion/\Ainter/\Adiff(set_1, set_2)$\\
		A10/A11&          $set \rightarrow \Agreater/\Aless(set, p, num)$\\
		A12&          $set \rightarrow \Aequal(set, p, num)$\\
		A13/A14&          $set \rightarrow \Aargmax/\Aargmin(set, p)$\\
		A15&          $set \rightarrow \Aset(e)$    \\		
		A16/A17/A18&          $e/p/num \rightarrow constant$ $^*$\\	\hline
	\end{tabular}
	\caption{\small Grammars to compose logical form. $^*$Instantiation for entity $e$, predicate $p$ or number $num$ from an input question.}
	\label{tab:csqa_grammer}
\end{table}

\subsection{Grammar and Logical Form}
We leverage similar formats of grammar and logical form as in \cite{d2a}. Here we give a brief introduction and refer readers to \cite{d2a} for more details.

\paragraph{Grammar}
The grammar definitions are shown in Table \ref{tab:csqa_grammer}, where each operator is composed of three parts, i.e., a semantic category, a function symbol and a list of arguments. An argument can be a semantic category or a constant instantiated from a question.  

\paragraph{Logical Form} 

A KB-executable logical form is usually formatted as a tree structure, where the root is the $start$ operator and each child node is a legitimate operator constrained by a semantic category in its parent's argument list. 
To take advantage of sophisticated sequence-to-sequence models \cite{bahdanau2015neural,vaswani2017attention} for question-to-logical-form translation, we re-format tree structure into a sequence by applying depth-first traversal over the tree. Reversely, once a sequence-formatted logical form is generated during decoding phase, it can be easily recovered into tree structure under grammars' guidance.

\subsection{Overview of Our Approach} \label{sec:proposed_alg}

\begin{figure}[t] 
	\centering
	\includegraphics[width=0.46\textwidth]{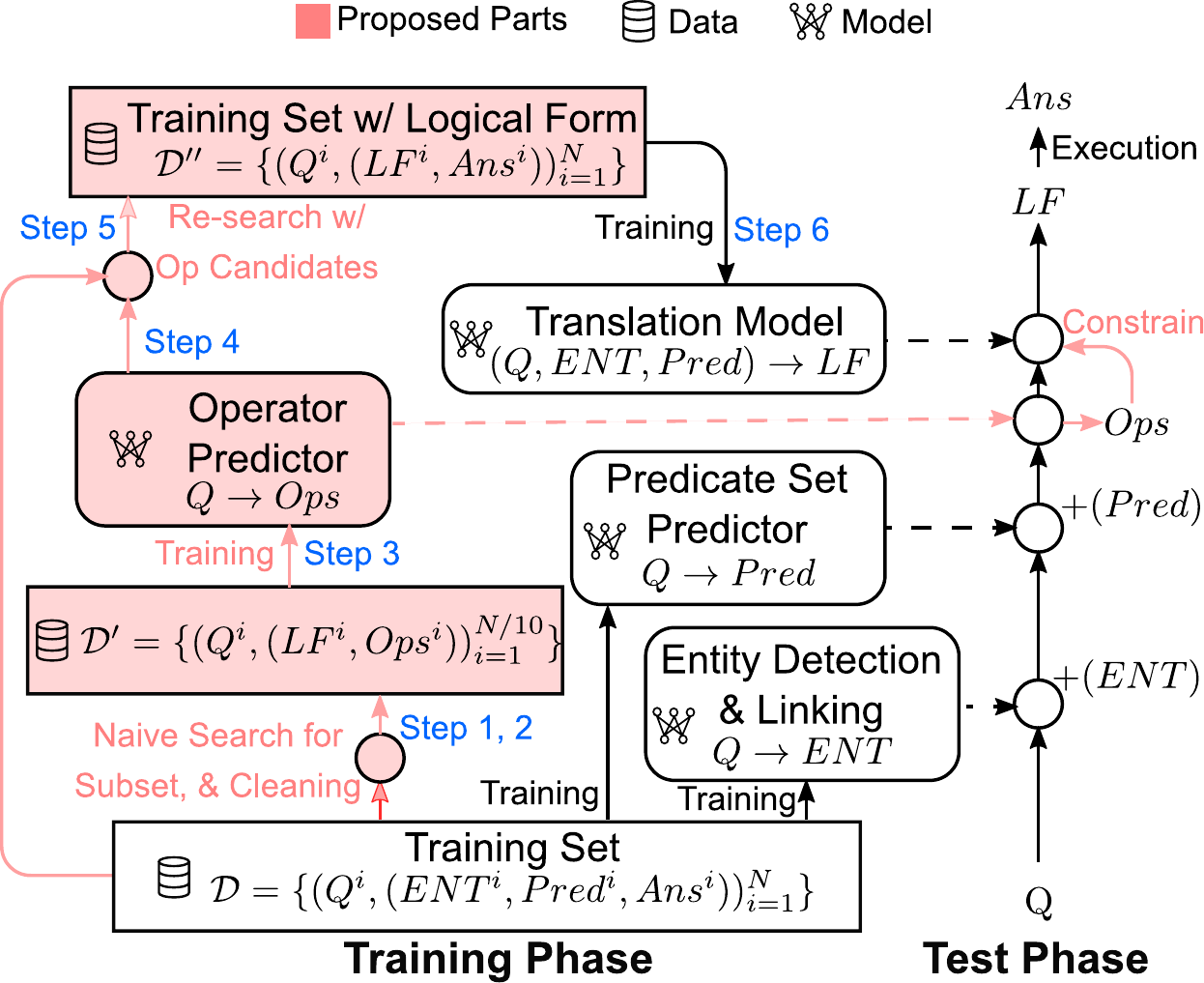}
	\caption{\small An overview of the proposed approach. }
	\label{fig:pipeline} 
	\centering
\end{figure}

As shown in Figure \ref{fig:pipeline}, our approach mainly consists of 6 steps: searching, cleaning, training operator predictor, operator prediction, re-searching, and training semantic parser.

\subsubsection{Searching and Cleaning}
Step 1 and 2 create training examples for proposed operator predictor based on a sampled small subset (e.g., $1/10$ of the total in this work) of the entire training data $\mathcal{D}$, which results in a new training set $\mathcal{D}'$. 

In Step 1, for each question $Q^i$ in the sampled small dataset, we naively search and get valid logical forms $LF^i = \{lf^{i1}, lf^{i2}, \dots\}$. As stated in Section \ref{sec:introduction}, this step could generate spurious logical forms and leads to bad operator predictor if we directly use these data for the model training. Hence, we further clean the searched results in Step 2. 

In Step 2, we clean searched logical forms according to question types, which is inspired by an observation that questions belong to same types require similar operators. 
For example, all these quantitative questions ``\textit{\underline{How many} cities are sister town of ...}", ``\textit{\underline{How many} rivers flow through ...}", and ``\textit{\underline{How many} countries have ...}" require the \textsc{Count} operator, rather than \textsc{Argmax}. 
We follow the same question types as \citeauthor{csqa}~\shortcite{csqa}, which is a general and widely-used taxonomy for KBQA. Specifically, we first create a legitimate operator set for each question type. The criterion is that, for a question type, an operator is legitimate if removing the operator from the candidates leads to a notable  (e.g., 1\% in our setup) search success ratio drop for the questions in that type. 
Then for each $LF^i$, we remove illegal logical forms that contain any operators not belonging to the legitimate set of corresponding question type. And, $Ops^i$ is a set of unique operators appearing in the cleaned $LF^i$. 

\subsubsection{Model Training for Operator Predictor}
In Step 3, an operator predictor model $\mathcal{M}$, which maps a question $Q^i$ into its most likely operators $Ops^i$ to compose a correct logical form, is trained based on the cleaned training data $\mathcal{D}'$. More details for operator predictor are introduced in Section \ref{sec:model_details}. 

\subsubsection{Training Data Generation for KBQA}
In Step 4 and 5, we apply $\mathcal{M}$ to each question in all training data $\mathcal{D}$ for predicting its most likely operators $Ops^i$, and re-search for valid logical forms $LF^i$ with constraints of $Ops^i$.  

\subsubsection{Model Training for KBQA}
Lastly, in Step 6, we train a semantic parser based on the searched results in Step 5. More details of the model are introduced in Section \ref{sec:model_details}.

Notably, this approach involves two rounds of searches, but it is still faster than previous works (e.g., naive BFS by  \citeauthor{d2a}~\shortcite{d2a}). The reason is that, in Step 1 only a small subset (e.g., 10\%) needs to be fully searched as previous works do, and with reduced search space, the re-search in Step 5 is much faster than previous works. For example, on CSQA benchmark, our algorithm is $\sim\!\! 3 \times$ faster than its baseline (search speed of 0.94s vs. 2.75s per example). 

There are two main benefits to incorporating an operator predictor into a standard ``\textit{searching} and \textit{training}'' scheme. 
First, it helps to provide high-quality data for downstream model training by improving search success ratio and reducing the number of spurious logical forms. Second, it makes training and inference more effective by providing the constraint from legal operators.

\subsection{Model Details} \label{sec:model_details}

We detailedly describe the implementation of our models in this section, including the operator predictor introduced in Step 3,  and semantic parsing model in Step 6 which consists of three sub-tasks: entity detection \& linking, predicate prediction, and sequence-to-sequence translation model. 

In formal terms, a question $Q$ is first tokenized as a list of words, i.e., $Q = [w_1, \dots, w_n]$, and then, a word embedding approach \cite{mikolov2013distributed} is invoked to transform the discrete words into low-dimensional vector representations, i.e., $X = [\bm{x_1}, \dots, \bm{x_n}]\in\mathbb{R}^{d_e \times n}$, where $d_e$ denotes embedding size and $n$ stands for question sequence length. The embedded words are separately passed into each following neural component with untied parameters. 

\paragraph{Operator Predictor} We define this sub-task as a multi-label problem, whose input is a natural language question and output is a set of operators possibly composing correct logical form for the question. In particular, a bi-directional LSTM ($\bilstm$) \cite{hochreiter1997long} performs over input word embeddings as an encoder to capture contextual information, which is denoted as 
\begin{align}
\bm{\overrightarrow{h}^{(s)}_{i}} &= \overrightarrow{\lstm}(\bm{x_i}, \bm{\overrightarrow{h}^{(s)}_{i-1}}; \theta^{(s1)}),~\forall i = 1, \dots, n, \label{eq:bilstm1_fw} \\
\bm{\overleftarrow{h}^{(s)}_{i}} &= \overleftarrow{\lstm}(\bm{x_i}, \bm{\overleftarrow{h}^{(s)}_{i+1}}; \theta^{(s2)}),~\forall i = n, \dots, 1, \label{eq:bilstm1_bw}\\
\bm{u^{(s)}} &= [\bm{\overrightarrow{h}^{(s)}_{n}}; \bm{\overleftarrow{h}^{(s)}_{1}}] \in\mathbb{R}^{2d_h}, \label{eq:bilstm1_cat}
\end{align}
where $\theta^{(*)}$ are learnable parameters for Bi-LSTM, $[;]$ denotes a concatenation operation, and $\bm{u^{(s)}}$ is the resulting vector representation for the whole question. Then, the probability of generating each operator $\bm{p^{(s)}}$ is defined as
\begin{equation}
	\bm{p^{(s)}} = \sigmoid(\mlp(\bm{u^{(s)}}; \theta^{(s3)})) \in\mathbb{R}^{|\mathcal{A}|},
\end{equation}
where, $\mlp(\cdot)$ is a multi-layer perceptron, $\mathcal{A}$ stands for all operators defined in Table \ref{tab:csqa_grammer}, and $|\mathcal{A}|$ denotes the size of $\mathcal{A}$. 



\paragraph{Entity Detection \& Linking} Entity detection aims to locate named entity mention in the input question, which is usually formulated as a sequence labeling problem. It assigns each word with one of \textit{B}, \textit{I} and \text{O} label, which stand for begin, middle and end of a named entity respectively. To solve this problem, we use  Bi-LSTM Conditional Random Field (Bi-LSTM-CRF) \cite{huang2015bidirectional} model to predict entity mention tag for each input word. Formally, another Bi-LSTM model is leveraged as a context embedding layer and it is parameter-untied with the one defined in Eq.(\ref{eq:bilstm1_fw})-(\ref{eq:bilstm1_bw}), i.e., 
\begin{equation}
	\bm{H^{(d)}} = \bilstm(\bm{X}; \theta^{(d1)}, \theta^{(d2)}),
\end{equation}
where, $\bm{H^{(d)}} = [\bm{h^{(d)}_1}, \dots, \bm{h^{(d)}_n}] \in\mathbb{R}^{2d_h \times n}$ and $\forall \bm{h^{(d)}_i} = [\bm{\overrightarrow{h}^{(d)}_{i}}; \bm{\overleftarrow{h}^{(d)}_{i}}]$. Then, a position-wise feed forward network with 3-way output is applied to each $\bm{h^{(d)}_i}$ for $i$-th word's scores over \textit{B}, \textit{I} and \text{O} respectively, which is written as
\begin{equation}
	\bm{S^{(d)}}  = \ffn(\bm{H^{(d)}}; \theta^{(d3)}),
\end{equation}
where, $\bm{S^{(d)}} \in\mathbb{R}^{3 \times n}$ denotes the resulting scores for all words, and $\theta^{(d3)}$ stands for the learnable parameters (i.e. weights and biases) for $\ffn(\cdot)$. Then, given a path of tags $\bm{l^{(d)}} = [l^{(d)}_1, \dots, l^{(d)}_n]~\text{where}~\forall l^{(d)}_i \in \{1,2,3\}$, a scoring function with learnable transition matrix $\bm{A}$ is defined as
\begin{equation}
	\score \nolimits(\bm{S^{(d)}}, \bm{l^{(d)}};  \bm{A}) \triangleq \sum_{i=1}^{n} (\bm{A}_{l^{(d)}_{i-1}, l^{(d)}_i} + \bm{S^{(d)}}_{l^{(d)}_i, i} ).
\end{equation} 
Further details on the training and inference of Bi-LSTM-CRF model are available in \cite{huang2015bidirectional}. 

Given mentions detected in the question, we then follow the traditional approach for linking them back to entities in a KB. 
Specifically, we first build an inverted dictionary where the keys are entity mentions and the values are linked entities with matching scores appended. Then, given a mention, we select the entity with highest score from the dictionary. 

\paragraph{Predicate Prediction} Identifying predicates in a question is also essential to compose an executable logical form. For this purpose, we simply formulate this sub-task as a multi-class prediction problem. 
In brief, another Bi-LSTM is used to embed the sentence as a vector representation, i.e., $\bm{u^{(p)}} = [\bm{\overrightarrow{h}^{(p)}_{n}}; \bm{\overleftarrow{h}^{(p)}_{1}}] \in\mathbb{R}^{2d_h}$ where $\bm{\overrightarrow{h}^{(p)}_{n}}$ and $\bm{\overleftarrow{h}^{(p)}_{1}}$ are derived from another $\bilstm(\cdot; \theta^{(p1)}, \theta^{(p2)})$. Then an $\mlp(\cdot; \theta^{(p3)})$ with $|\mathcal{P}|$-way $\softmax$ is used to fulfill the classification, where $\mathcal{P}$ is a set of all possible predicates. In training phase, a negative log-likelihood loss is applied to learn this model's parameters. 

\paragraph{Question-to-Logical-Form Translation} Given predicted entity and predicate candidates from the upstream, a semantic parsing model aims to translate a input natural language question into KB-executable logical form. Since the logical forms have been formatted as sequences, we employ a sequence-to-sequence encoder-decoder structure with attention mechanism \cite{bahdanau2015neural}. In particular, we here use a Bi-LSTM model as an encoder for natural language question to produce context-aware representation for each word, which is formulated as
\begin{equation}
\bm{H^{(t)}} = \bilstm(\bm{X}; \theta^{(t1)}, \theta^{(t2)}),
\end{equation}
where, $\bm{H^{(t)}} = [\bm{h^{(t)}_1}, \dots, \bm{h^{(t)}_n}] \in\mathbb{R}^{2d_h \times n}$ and $\forall \bm{h^{(t)}_i} = [\bm{\overrightarrow{h}^{(t)}_{i}}; \bm{\overleftarrow{h}^{(t)}_{i}}] \in\mathbb{R}^{2d_h}$. For the decoder, we employ a forward $\lstm$ as an autoregressive model to predict logical form. To be specific, at $j$-th decoding step, given previous hidden state $\bm{s_{j-1}}$, we use a compatibility function to calculate the alignment score between previous decoding hidden state and each encoding  word representation, resulting in a contextual embedding $\bm{c_j}$. The attention procedure and the decoder's hidden state update formula are expressed as
\begin{align}
	\bm{\alpha _j} &= \softmax(\bm{s_{j-1}}^T \bm{W^{(t)}} \bm{H^{(t)}}) ^T,\\
	\bm{c _j} &= \bm{H^{(t)}} \bm{\alpha _j}, \\
	\bm{s_{j}} &= \lstm([\bm{v _j}; \bm{c _j}], \bm{s_{j-1}}; \theta^{(t3)}),~\forall j = 1, \dots, m
\end{align}
where $\bm{W^{(t)}} \in\mathbb{R}^{d_h \times 2d_h}$ is a learnable parameter matrix, $\bm{\alpha _j} \in\mathbb{R}^{n}$ is attention distribution over all encoder states, $\bm{v _j}$ is decoding input embedding for $j$-th step. Next, given the decoding hidden state $\bm{s_{j}}$, a neural classifier composed of a linear layer is used to predict an operator for $j$-th step, i.e., 
\begin{align}
	\bm{\hat{y}_j} = \softmax(\ffn(\bm{s_j }; \theta^{(t4)})) \in\mathbb{R}^{|\mathcal{A}|},
\end{align}
where $\bm{\hat{y}_j}$ is a predicted distribution over all possible operators, i.e., $\mathcal{A}$. 
Unlikely typical sequence-to-sequence task, e.g., neural machine translation, except predicting one operator from Table \ref{tab:csqa_grammer} at each step, the model also needs to instantiate a semantic category. Specifically, if predicted operator is one of $A16$, $A17$ or $A18$ (i.e., entity, predicate, number), the model is also required to choose one term from corresponding candidates for instantiation. Note, number candidates are derived from named entity recognition by SpaCy. 

To complete instantiation, entity(s), predicate(s) and number(s) are respectively embedded by mean-pooling over entity's composing words, predicate token embedding, and character-level 1D-CNN \cite{kim2014convolutional}. Then, a dot-product is invoked between decoding hidden state $\bm{s_{j}}$ and candidates' embeddings of targeted instantiation type (w.r.t. the predicted operator). A $\softmax$ is finally applied to all scores of the candidates to produce a prediction distribution. After iterative decoding with grammar's guidance, a logical form can be completely composed by this autoregressive model. 

Ideally, all four models introduced above would be trained in a multi-task learning framework to make the best of the shared encoder and improve the performance. However, our focus in this paper is to propose an effective and efficient approach to search full supervision data for weakly supervised KBQA, and thus highlight that more high-quality training data play a vitally important role in this task. Hence, for a fair comparison, we directly adopt previous state-of-the-art pipeline model \cite{d2a} in our framework, rather than attempting to improving it. In addition, ``copy'' operators proposed by \citeauthor{d2a}~\shortcite{d2a} are also included in this model for competitive results, whose descriptions are omitted for simplification.

\section{Experiment}
\begin{table*}[t] 
	\centering
	\begin{tabular}{l|c|cc|cc|cc}
		\hline
		\textbf{Methods} &    & \multicolumn{2}{c|}{HRED+KVmem\textbf}      & \multicolumn{2}{c|}{D2A (Baseline)}     & \multicolumn{2}{c}{\textbf{D2A+Ours}*}      \\ \hline
		\textbf{Question Type}  & \#Example     & Recall       & Precision      & Recall       & Precision      & Recall       & Precision     \\ \hline
		Overall &    -  &  18.40\%        &  6.30\%    & 66.83\% & 66.57\% &  \textbf{71.63}\%             &  \textbf{72.42}\%             \\
		Simple Question (Direct)     & 82k    &  33.30\%          &  8.58\%                 & 79.50\% & 77.37\% &  \textbf{82.80}\%             &  \textbf{83.20}\%             \\
		Simple Question (Coreferenced)    & 55k    &  12.67\%          &  5.09\%                 & 58.47\% & 56.94\% &  \textbf{64.67}\%             &  \textbf{64.58}\%             \\
		Simple Question (Ellipsis)   & 55k     &  17.30\%          &  6.98\%                 & 84.67\% & 77.90\% &  \textbf{84.88}\%             &  \textbf{83.02}\%             \\
		Logical Reasoning (All)   & 22k     &  15.11\%          &  5.75\%                 & 65.82\% & 68.86\% &  \textbf{73.88}\%             &  \textbf{72.00}\%             \\
		Quantitative Reasoning (All)    & 9k     &  0.91\%          &  1.01\%                & 52.74\% & 60.63\% &  \textbf{60.30}\%          &   \textbf{68.06}\%             \\
		Comparative Reasoning (All)    & 15k      &  2.11\%          &  4.97\%              & 44.14\% & 54.68\% &  \textbf{50.42}\%            & \textbf{60.62}\%             \\
		Clarification    & 12k     &  25.09\%          &  12.13\%                & 37.24\% & 33.97\% &  \textbf{38.74}\%            &  \textbf{34.80}\%           \\ \hline
		\textbf{Question Type}   &\#Example     & \multicolumn{2}{c|}{Accuracy} & \multicolumn{2}{c|}{Accuracy} & \multicolumn{2}{c}{Accuracy} \\ \hline
		Verification (Boolean)    &27k     & \multicolumn{2}{c|}{21.04\%} &  \multicolumn{2}{c|}{37.07\%}        & \multicolumn{2}{c}{\textbf{45.80\%}}          \\
		Quantitative Reasoning (Count)    &24k   & \multicolumn{2}{c|}{12.13\%} &  \multicolumn{2}{c|}{38.42\%}  & \multicolumn{2}{c}{\textbf{41.35\%}}           \\
		Comparative Reasoning (Count)   &15k  & \multicolumn{2}{c|}{8.67\%} &  \multicolumn{2}{c|}{16.62\%}  & \multicolumn{2}{c}{\textbf{20.93\%}}          \\ \hline
	\end{tabular}
	\caption{\small KBQA result comparison with HRED+KVmem  \cite{serban2016building,miller2016key} and D2A \cite{d2a}. *D2A+Ours means that D2A model is integrated with the proposed pipeline as shown in Figure \ref{fig:pipeline}.}
	\label{tb:csqa_res} 
\end{table*}

This section begins with experimental setups to evaluate our proposed framework. Then, the evaluation includes assessments of the quality of responses to a KBQA task, effectiveness of searching, and the performance of each sub-task. Lastly, case study and error analysis are presented for qualitative and in-depth understanding of this work. 
\subsection{Experimental Settings}
\paragraph{Dataset} 
We employed one of the largest weakly-supervised KBQA datasets over large-scale knowledge base, Complex Sequential Question Answering (CSQA) \cite{csqa} in our experiments. There are 1.6M turns in 200K dialogues without logical form labeled. Its KB is built on \textsc{WikiData} in the form of (\textsl{subject}, \textsl{predicate}, \textsl{object}), including 21.2M triplets over 12.8M entities. Moreover, it also defines a question taxonomy with 10 types (e.g., logical reasoning and comparative reasoning), and labels each question in the dataset with the type it belongs to. Although the taxonomy is defined for this dataset, it is rather general and can be used to other KBQA datasets.

\paragraph{Evaluation Metrics} 
In line with \citeauthor{csqa}~\shortcite{csqa} and \citeauthor{d2a}~\shortcite{d2a}, we used \textit{Precision} and \textit{Recall} as metrics for questions when the answer is an entity(s), and \textit{Accuracy} for questions when the answer is  boolean or numeric.

\paragraph{Model Setup} For each neural model, the word embedding weight matrix was independent of each other and the embedding size $d_e$ was 300D; the hidden state size $d_h$ was also set to 300D and activation function was $\relu$ for the middle layer of each $\mlp$. For the optimization, we used Adam optimizer \cite{kingma2014adam} with learning rate of $10^{-3}$, the batch size was set to 64 for 6 epochs, and early stop strategy was applied when there was no longer a significant improvement over the development set during the training. Moreover, for the operator predictor we first used naive BFS method to search only 10\% training data from CSQA and applied the data pre-processing steps outlined in Section \ref{sec:proposed_alg}.

\paragraph{Baselines} Only a few approaches have been proposed for solving large-scale, weakly supervised KBQA problem. HRED+KVmem \cite{csqa} and D2A \cite{d2a} are two typical approaches in regard to information retrieval and neural symbolic ways, respectively. In particular, HRED+KVmem involves a sequence-to-sequence based HRED model \cite{serban2016building} and a key-value memory network \cite{miller2016key} to retrieve answer from KB. 
In contrast, D2A\footnote{The re-implemented D2A in this work outperforms the one originally proposed by \citeauthor{d2a}~\shortcite{d2a}, and one possible reason is that our re-implemented grammars reach a better performance balance between simple and non-simple questions. For a fair comparison, we report the re-produced results for D2A in this paper.} defines a set of semantic parsing grammars and translates natural language questions into corresponding logical forms to query KB via a memory-augmented neural symbolic model.

\subsection{Question Answering Performance}\label{sec:exp_qa}

As listed in Table \ref{tb:csqa_res}, our proposed effective search approach coupled with D2A model improves previous baselines by a significant margin, setting a new state-of-the-art performance on CSQA dataset. 
Specifically, compared to the strong baseline, D2A, our proposed framework can improve the recall from 66.83\% to 71.63\% and the precision from 66.57\% to 72.42\% in terms of the overall score. 
And, as shown in the bottom panel of Table \ref{tb:csqa_res}, for boolean (i.e, \textit{Verification}) and numeric (i.e., \textit{Quantitative} and \textit{Comparative Reasoning}) questions, our proposed framework also significantly outperforms D2A. 
In addition, the improvements are more notable with more complex question types. For example, 7.56\%/7.43\% improvement of recall/precision over \textit{Quantitative Reasoning} is much greater than 3.30\%/5.83\% improvement over \textit{Simple}.
We attribute this to more operators required to answer more complex questions, which exacerbates the problems associated with large search space.

\subsection{Searching Effectiveness}

In this section, we quantitatively analyze the effectiveness of our proposed algorithm in terms of alleviating the two problems caused by large search space, i.e., low search success ratio and spurious logical forms. 

\begin{figure}[t] %
	\centering
	\includegraphics[width=0.45\textwidth]{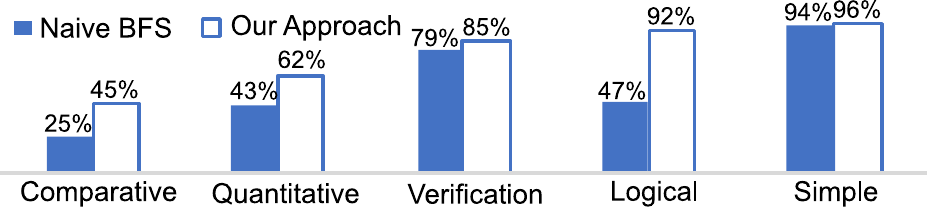}
	\caption{\small Search success ratio comparison w.r.t. question type.}
	\label{fig:suc_ratio} 
	\centering
\end{figure}

\begin{table}[t]\small 
	\centering
	\begin{tabular}{l|l|l|l}
		\hline
		Method    & \#Correct & \#Spurious & \%Spurious \\ \hline
		Naive BFS & 114       & 136        & 54.5\%     \\ 
		\textbf{Ours}      & \textbf{115}       & \textbf{40}         & \textbf{26.7\% }    \\ \hline
	\end{tabular}
	\caption{\small Statistics of spurious logical forms. Note that a question may be assigned with multiple valid logical forms.}
	\label{tb:spu}
\end{table}

\paragraph{Increasing Search Success Ratio}
The search success ratio is defined as the number of questions, each with at least one valid logical form found by a search method, over the total number of questions. We compared our approach with traditional BFS one \cite{d2a}, and reported the results in Figure \ref{fig:suc_ratio}. From the figure we found that our approach increases the search success ratio significantly, especially almost $2\times$ increase for \textit{logical reasoning} questions. And we also found, the improvement of KBQA is proportional to the increase of search success ratio w.r.t. question types.

\paragraph{Reducing Spurious Logical Forms}
To determine whether our approach reduces the number of spurious logical forms, we randomly sampled 40 questions, each with at least one valid logical form found through both naive BFS search method and our approach. Human evaluators manually inspected the results and made a judgment as to whether the logical form is correct or spurious. The results listed in Table \ref{tb:spu} demonstrate that our approach considerably reduces the incidence of spurious logical forms from 54.5\% to 26.7\% compared to the baseline. This is a substantial reduction and provides a clear evidence that this approach can improve the quality of training data and thus benefit any downstream semantic parsing model.

\subsection{Sub-Task Evaluation}

We evaluated performance of the models for solving sub-tasks, which provide prerequisites for question-to-logical-form translation model. These empirical results can also measure the severity of error propagation in pipeline model. 

\subsubsection{Operator Predictor}

\begin{table}[t]\small 
	\centering
	\setlength{\tabcolsep}{3pt}
	\begin{tabular}{l|c|c}
		\hline
		\textbf{Question Type}    & \textbf{Question Coverage (\%)} &\textbf{Remaining (\%)}\\ \hline   
		Quantitative Reas&	97.38  & 44.15\\  
		Comparative Reas&	94.41 &	 50.93\\  
		Verification &	99.06 &	  17.65\\   
		Logical Reasoning &	98.07	&  23.90 \\  
		Clarification &	86.21  &	  21.50 \\  
		ALL &	98.67	&30.83     \\ \hline  
	\end{tabular}
	\caption{\small Operator predictor evaluation.}
	\label{tb:ops}
\end{table}

A proper operator prediction is crucial to downstream semantic parsing model training. A poorly-trained operator predictor will lead to searching in a wrong logical form space, and consequently damage the performance of the model. 
To assess the quality of the operator predictor, we took (question, valid operators) pairs found by naive BFS as evaluation set, and evaluated the performance according to a metric, i.e., \textit{Question Coverage}. 
\textit{Question coverage} is defined as the number of questions, with predicted operators able to compose at least one valid logical form, as a ratio of all questions. 
As shown in Table \ref{tb:ops}, our operator predictor can achieve 98.67\% of question coverage, which means when re-searching logical forms in Step 5, our approach will locate a correct sub-space for at least 98.67\% questions.  
As an auxiliary measure, \textit{Remaining} represents averaged proportion of the size of predicted operator candidates over the number of all operators defined in Table \ref{tab:csqa_grammer}. 



\begin{table*}[htbp]\small
	\centering
	\begin{tabular}{p{3.6cm}|p{5.6cm}|p{4.4cm}p{2.4cm}}
		\toprule
		\textbf{Question}&  \textbf{Logical Form from Naive Approach}&  \textbf{Logical Form from Ours}&  \textbf{Ops Prediction}\\
		\midrule
		Where is Zinc finger protein 775 found?&  
		\textsc{Diff(Find(Set(\textit{Zinc...775}, \textsl{{found-in-taxon}})), Set(\textit{Zinc...775}))}&   
		\textsc{Find(Set(\textit{Zinc...775}, \textsl{{found-in-taxon}}))}$^{*}$&  
		[\textit{start $\rightarrow$ set}, \textsc{Find}, \textsc{Set}]\\
		\midrule
		Is Sumy Oblast adjacent to Poltava Oblast?&  
		\textsc{In(\textit{Poltava Oblast}, Union(Find(Set( \textit{Sumy Oblast}), \textsl{{shares-border}}), Set(\textit{Italy})))}&  
		\textsc{In(\textit{Sumy Oblast}, Find(Set( \textit{Poltava Oblast}), \textsl{{shares-border}}))}$^{*}$&  
		[\textit{start $\rightarrow$ bool}, \textsc{In}, \textsc{Find}, \textsc{Set}]\\ 
		\midrule
		Which administrative territories holds diplomatic relationship with max number of administrative territories?&  
		\textsc{Diff(Argmax(Count(Find(Find(Set( \textit{administrative territorial}), \textit{is-a})), \textsl{{diplomatic-relation}})), Set(\textit{Quebec}))}&  
		\textsc{Argmax(Count(Find(Find( Set(\textit{administrative territorial}), \textsl{{is-a}})), \textsl{{diplomatic-relation}}))}$^{*}$&  
		[\textit{start $\rightarrow$ set}, \textsc{Find}, \textsc{Count}, \textsc{Union}, \textsc{Diff}, \textsc{Argmax}, \textsc{Set}]\\ 
		\midrule
		Which administrative territories are Yale University present in and are the origins of Anna Karenina?&
		\textsc{Inter(Find(Set(\textit{Yale University}), \textsl{{country}}), Find(Set(\textit{Anna Karenina}), \textsl{{country-of-origin}}))}$^{*}$& 
		\textsc{Find(Set(\textit{Yale University}, \textsl{{country}}))}& 
		[\textit{start $\rightarrow$ set}, \textsc{Find}, \textsc{Count}, \textsc{Union}, \textsc{Inter}, \textsc{Diff}, \textsc{Set}]\\
		\bottomrule
	\end{tabular}
	\caption{Case study of valid logical forms. And a logical form ending with $^{*}$ means it is correct, otherwise spurious.}
	\label{tb:case_study}
\end{table*}

\subsubsection{Entity Detection \& Linking} The employed entity detection model is quite accurate when predicting ``IOB'' tags for the entity mentions. The F1 score of this sequence labeling task can reach $\sim$99\%. 
Given detected entity mentions, we need to link them back to the knowledge base, so we also evaluated the performance of entity linking, and obtained\footnote{Co-references may appear in the example on CSQA and thus the oracle linking label is usually inaccurate, so our entity detection \& linking model is underestimated. } the precision of 24\% and the recall of 90\%. 
The results show that, although the precision is relatively low due to entity ambiguity, the recall stays high, indicating that most correct entities are retrieved and sent to downstream for the translation. The low precision mainly comes from ambiguity of entities, which means different entities share the same mention but express totally different meanings. 

\subsubsection{Predicate Prediction}
As demonstrated in Section \ref{sec:model_details}, the predicate classification is formulated as a multi-class problem in training phase. During the inference, in case of more than one predicates existing in an input question, we kept the top-$n$ most probable predicates. Although this certainly hurts the precision, it can guarantee a high recall and reduce error propagation. The final results are a precision of 46\% and a recall of 98\% with $n$ set to two.

\subsection{Case Study} 

In this section, we leverage some cases to demonstrate the effectiveness of our proposed algorithm in searching for logical forms on weakly supervised KBQA. As shown in Table \ref{tb:case_study}, for each question, we listed the logical forms searched by naive BFS approach and our proposed one respectively, as well as the predicted operators from \textit{operator predictor}. 

According to first three cases, due to the constraints posted by \textit{operator predictor} (last column), our approach could avoid some spurious results. Meanwhile, as shown in fourth case, although predicted operator candidates substantially reduces the search space, it is still possible to include spurious logical forms in searched results.

\subsection{Error Analysis} 
To conduct an error analysis and provide an insight into the causes of the prediction errors, we randomly sampled 50 wrongly-predicted examples for KBQA, and found the errors could be coarsely categorized as follows. 

\paragraph{Entity Ambiguity} 
This is the most serious problem leading to wrong predictions during question-to-logical-form translation since many entities with identical text however express totally different meanings. For example, an entity \textit{The Avengers} could be a movie, a soundtrack album or a punk rock band; even for a movie whose title is \textit{The Avengers}, it also could be \textit{2012 superhero film produced by Marvel} or \textit{1998 film by Jeremiah S. Chechik}. 

\paragraph{Error Propagation}
Because a pipeline approach is employed to solve KBQA problem, it is inevitable that the prediction errors occurring at early stage will be propagated into downstream models. An apparent case is that wrongly-predicted predicate candidates directly lead to an un-executable logical form. 

\paragraph{Translation Error} Due to translation model's limitation on representative expression, a wrong operator or entity could be chosen to compose a logical form during decoding, which results in an incorrect answer. 

\section{Related work}

This work is in line with semantic parsing based approach for KBQA task. Given a natural language question, based on a set of well-defined grammars for specific task, typical semantic parsing approaches learn a model to transform the question to a KB-executable logical form for answer retrieval \cite{wong2007learning,zettlemoyer2009learning,kwiatkowski2011lexical,andreas2013semantic,artzi2013weakly,zhao2014type,long2016simpler,jia2016recombination,ling2016latent,xiao2016sequence}. 

Usually, because of limited crowdsourcing, only final answers instead of full executable logical forms are provided to learn a semantic parsing model, i.e., in a weakly supervised learning scheme \cite{webquestions,iyyer2017search,csqa}. 
Hence, ``\textit{searching} and \textit{training}'' is a conventional stepwise approach to handle such weakly supervised setting by searching logical form for semantic parser learning \cite{kbqaasmt,stagg,macrogrammar,ltop,mapo,d2a,dasigi2019iterative}. 

However, \textit{searching} over structured knowledge bases inevitably leads to spurious logical form problem, which introduces wrongly labeled data and thus poses negative effect on KBQA performance \cite{pasupat2016inferring,d2a}. To alleviate spurious logical forms' effect, for example, \citeauthor{mapo}~\shortcite{mapo} separately estimated expectations over the trajectories inside and outside high-rewarded memory buffer, rather than maximum likelihood training. 
\citeauthor{ltop}~\shortcite{ltop} reduced the impact of spurious logical forms by using randomized beam search and more balanced optimization. 
And, \citeauthor{dasigi2019iterative}~\shortcite{dasigi2019iterative} alternated between searching for consistent logical forms and maximizing the marginal likelihood of the retrieved ones while iterative training, which increases logical forms' complexity for subsequent ones, thus dealing with the problem of spuriousness.  
In addition, there are also some works proposed to reduce the search space. \citeauthor{macrogrammar}~\shortcite{macrogrammar} used macro grammars to reduce the search space. 
And, \citeauthor{dong2018coarse}~\shortcite{dong2018coarse} proposed coarse-to-fine semantic parsing model to predict logical form sketch first. 

In contrast, our approach aims to prevent the problems from its root. In other words, we directly reduce search space by restricting operator candidates, which decreases spurious logical forms in search results, and also increases search success ratio in the meantime. 

\section{Conclusion}
We proposed a novel approach for effective search of logical forms by operator prediction for weakly supervised KBQA task, which provides sufficient and superior data for downstream question-to-logical-form translation model training, and makes training and inference more effective under the constraints of  possible operators. The proposed approach is simple and effective, which makes it of great practical use. Experimental results verify the effectiveness of our approach in terms of reducing spurious logical forms, increasing search success ratio, improving search efficiency, and boosting the final accuracy for question answering.

{
\small
	\bibliography{ref}
	\bibliographystyle{aaai}
}

\end{document}